\documentclass[conference]{IEEEtran}
\IEEEoverridecommandlockouts
\usepackage{caption}
\usepackage{placeins}

\usepackage{lipsum}
\usepackage{cite}
\usepackage{tabularray}
\usepackage{threeparttable}
\usepackage{pdfpages}
\usepackage{booktabs}
\usepackage{multirow} 
\usepackage{float}
\usepackage{tabularx}
\usepackage{fixltx2e}
\usepackage{hyperref}
\usepackage{tikz}
\usetikzlibrary{calc,patterns,angles,quotes}
\usepackage{cancel}
\usepackage{graphicx}

\usepackage{amsmath,amssymb,amsfonts,amsthm}
\usepackage{algorithmic}
\usepackage{graphicx}
\graphicspath{ {./images/} }
\usepackage{subfigure}
\usepackage{multicol}
\usepackage[export]{adjustbox}
\usepackage{textcomp}
\usepackage{xcolor}
\usepackage{xfrac}    % for \xfrac macro
\usepackage{nicefrac} % for \nicefrac macro
\usepackage{amsmath}  % for \dfrac macro
\usepackage{mathtools}
\def\BibTeX{{\rm B\kern-.05em{\sc i\kern-.025em b}\kern-.08em
    T\kern-.1667em\lower.7ex\hbox{E}\kern-.125emX}}
    
\begin{document}

\title{QWID: Quantized Weed Identification Deep neural network}

\author
{
    Parikshit Singh Rathore \\ 
    \IEEEauthorblockA{\textit{Maharana Pratap University of Agriculture and Technology, India}}
    \href{mailto:14.parikshitsingh@gmail.com} {14.parikshitsingh@gmail.com}
    % \IEEEauthorblockN{Parikshit Singh Rathore\thanks{ Email: 14.parikshitsingh@gmail.com }}
}

\maketitle

\begin{abstract}
In this paper, we present an efficient solution for weed classification in agriculture. We focus on optimizing model performance at inference while respecting the constraints of the agricultural domain. We propose a Quantized Deep Neural Network model that classifies a dataset of 9 weed classes using 8-bit integer (int8) quantization, a departure from standard 32-bit floating point (fp32) models. Recognizing the hardware resource limitations in agriculture, our model balances model size, inference time, and accuracy, aligning with practical requirements. We evaluate the approach on ResNet-50 and InceptionV3 architectures, comparing their performance against their int8 quantized versions. Transfer learning and fine-tuning are applied using the DeepWeeds dataset. The results show staggering model size and inference time reductions while maintaining accuracy in real-world production scenarios like Desktop, Mobile and Raspberry Pi. Our work sheds light on a promising direction for efficient AI in agriculture, holding potential for broader applications. \footnote{GitHub: \url{https://github.com/parikshit14/QNN-for-weed}}

\end{abstract}

\section{Introduction}
Weeds are undesirable plants that compete with the agricultural crop plant for resources like soil nutrients, direct sunlight, water and to some extent space to grow. The weeding process plays a significant role in agriculture because weeds produce a major loss in crop yield. With as high as 50-71\% yield reduction was seen in soybean, to around 40-71\% in groundnut~\cite{GHARDE201812}. Weeds accounted for a total of 12.65 billion dollar losses in 2007 in India alone ~\cite{varshney2008future}.

Weed identification plays an important role in weeding because deriving the weed type with appropriate hoeing depth, hoeing positions, and particular herbicides can be used ~\cite{wang2022weed25}.
Automating these processes minimizes human intervention which is high in cost and also labor-intensive.

\begin{figure}[!htbp]
    \centering
    \includegraphics[width=\linewidth]{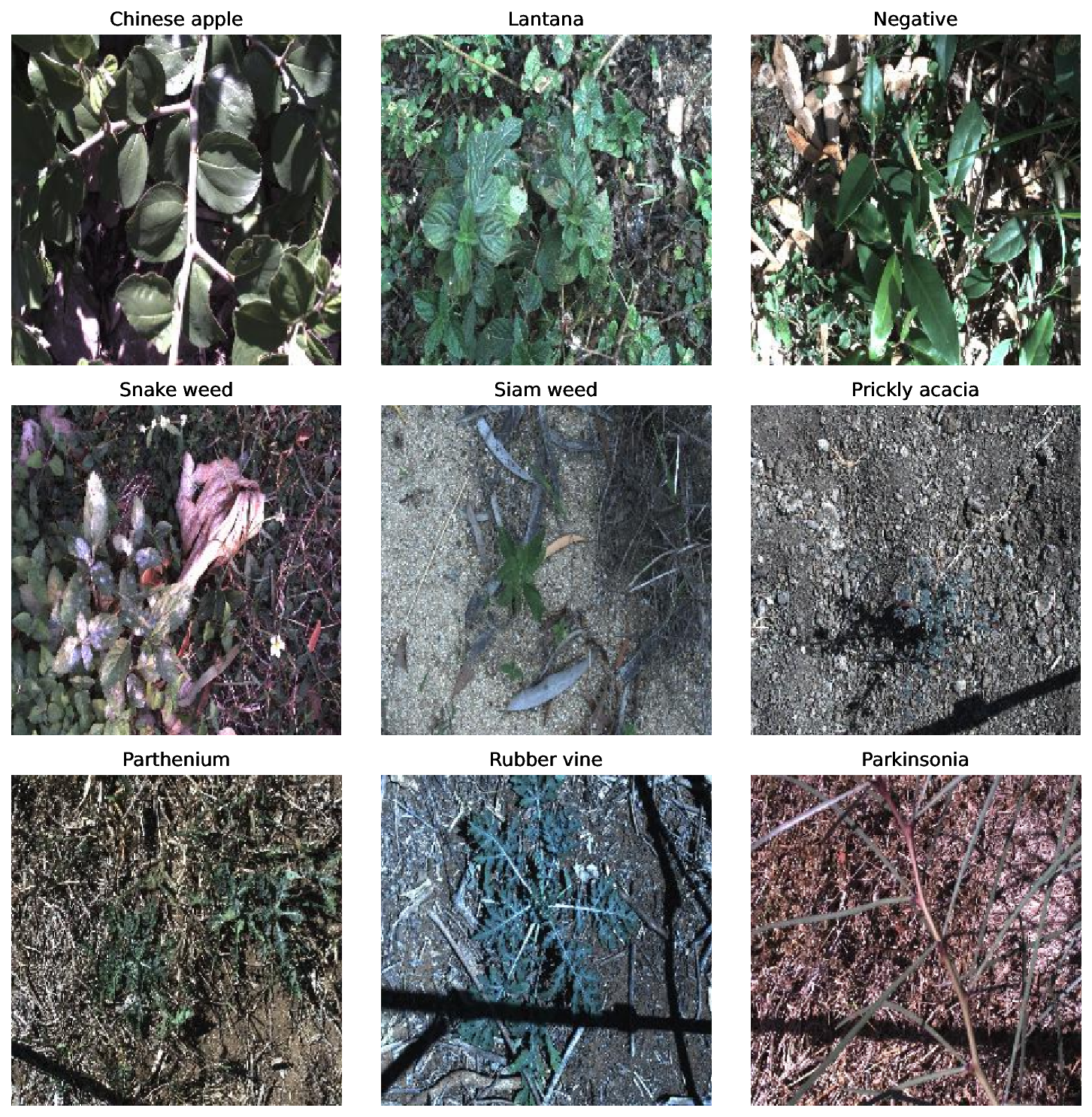}
    \caption{Dataset samples \cite{olsen2019deepweeds}}
    \label{fig:weedcollage}
\end{figure}

Training the models on DeepWeeds dataset \cite{olsen2019deepweeds} consisting of 9 classes namely chinese apple, lantana, parkinsonia, parthenium, prickly acacia, rubber vine, siam weed, snake weed and negatives (other non-target plant life). The dataset was prepared in real-world conditions like dark shadows, canopy cover, high contrast, and variable distance between the camera and the plant. Sample of each class is represented in Figure \ref{fig:weedcollage}.

Models used originally for this dataset were ResNet-50 \cite{he2015deep} and Inceptionv3 \cite{szegedy2016rethinking}, both transfer learned on the DeepWeeds dataset. We put forward the quantized versions of both ResNet-50 and Inceptionv3, transfer learned and fine-tuned to achieve almost the same accuracy but with significantly better inference time and model size. The int8 version is also better suited than the fp32 in production as it requires low computational power which is not readily available in farmlands.

The state-of-the-art (SOTA) object classification models aim to increase the model accuracy by building a denser neural network with precise calculations in terms of weights, biases, activation functions, matrix multiplications, etc., which leads to high accuracy. As a result, it increases the number of calculations which in turn is time-consuming even for inference (forward pass). These computationally heavy models require high-performance servers or workstations equipped with GPUs enabled for parallel computing. While on embedded devices, the computational resources present are very limited, an offline execution of the model takes place, i.e., the inference is performed on powerful servers instead of the target system. Another approach includes using lightweight deep neural network models like MobileNet \cite{howard2017mobilenets} which uses depth-wise separable convolutions to reduce the model complexity, resulting in architectural changes which has limited capacity for complex patterns. 
% the proposed resnet and inception model leverage the power of low complexity  
In this paper, we propose a quantized ResNet and quantized Inception weed classification model to address the limited computational resources on edge devices. Although training a quantized model leads to an accuracy drop from the SOTA models, the drop is very marginal, a mere 1-3\% while being able to achieve more than 10 times gain in performance, in terms of inference time when compared to a non-quantized model.

The remainder of the paper is organized as follows. In Section \ref{sec:intro}, we discuss the related research. In Section \ref{sec:training}, we present our approach, which includes the architecture, training methodology, and the associated issues with quantization. In Section \ref{sec:results}, we present our findings on the DeepWeeds dataset and share model results based on relative accuracy, inference time, and complexity, followed by a conclusion in Section \ref{sec:conclusion}.

\section{Related Work} \label{sec:intro}
\subsection{Quantization}
Although similar concepts had first appeared in the literature as early as 1898, the history of the theory and practice of quantization dates back to 1948. The early development of pulse code modulation systems led to the initial recognition of quantization in modulation and analog-to-digital conversion.

In neural networks, quantization is used to reduce the memory consumption of weight biases and activation by using low-precision datatypes like int8 instead of fp32. This reduces the model size by a factor of four. For context, the amount of multiplication and addition operations produced by operating a neural network on hardware can quickly reach many millions. High precision is typically not required during inference and could impede the use of AI in real-time or on devices with limited resources. Large computational gains and improved performance are obtained by combining lower-bit mathematical operations with quantized parameters for the intermediate calculations in a neural network.

\begin{figure}[!htbp]
    \centering
    \includegraphics[width=\linewidth]{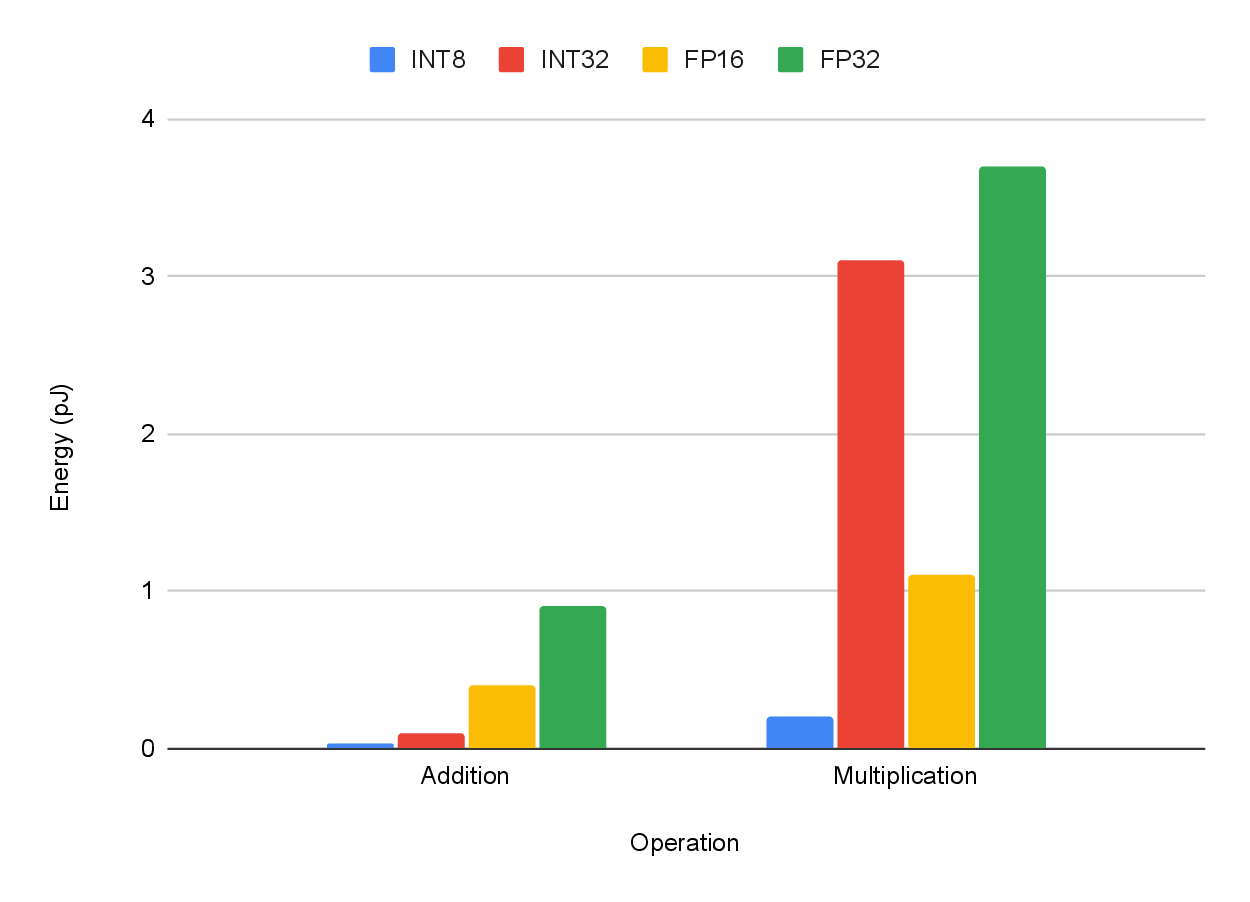}
    \caption{Energy consumption on 45nm processor \cite{6757323}}
    \label{fig:energy}
\end{figure}

Quantized neural networks improve power economy in addition to performance for two reasons, i.e., decreased memory access costs, and improved computation efficiency. By utilizing the lower-bit quantized data, less data must be moved both on and off-chip, reducing memory bandwidth and significantly reducing energy consumption. Mathematical operations with lower precision, such as an int8 multiplication as opposed to an fp32 multiplication, use less energy and have a higher compute efficiency, which results in less power being used. Sample comparison of power consumption in Figure \ref{fig:energy}.
\subsection{CNN-Based Image classification}
Several papers propose CNN-based models to improve the accuracy on the DeepWeeds dataset. The original paper on DeepWeeds \cite{olsen2019deepweeds} proposes a ResNet-50 and Inceptionv3 models trained using transfer learning with fine tuning on ImageNet weights,\cite{chen2022performance} presents 27 SOTA deep learning models through transfer learning evaluating accuracies and inference of each, \cite{huertas2022fusing} proposes a combination of predictions of a CNN and a secondary classifier for statistical features in weed images. \cite{chen2023deep} proposes a diffusion probabilistic model to generate synthetic weed images of high quality to overcome the cost of data capture along with transfer learning, \cite{Zhang_2023} proposes a combination of convolutional neural network and transformer structures for classification and feature extraction.
However, none of the papers propose an improvement in terms of inference time and low computational power as present in actual agricultural environments.

\subsection{CNN-Based Image classification for Embedded systems}
The SOTA CNN models require high-end GPU not only for training but also for the purpose of inference, else it increases the inference time drastically. As a result, best-performing CNN models fail to outperform optimally on embedded systems on chips (SoC) \cite{10.1145/2847263.2847265}. For instance, a provider in the security camera market discovered that even after switching the YOLO back-end from GoogleNet to a more straightforward CNN like AlexNet, their embedded implementation only operates at a maximum frame rate of 5 frames per second on embedded GPUs \cite{tripathi2017lcdet}.

In our proposed algorithm, a quantized convolutional neural network (Q-CNN), a form of compressed and accelerated CNN model, is implemented for the DeepWeeds dataset without significant accuracy loss. The proposed models use PyTorch \cite{paszke2019pytorch}, an open-source framework for model training and inference. It is heavily used in research and development tasks in industry and academia. The PyTorch quantization module currently provides support for x86 CPUs, ARM CPUs which are typically found in mobile/embedded devices, and early support for Nvidia GPU via TensorRT. On these systems, the suggested PyTorch-based model can benefit from similar architectural advantages and can be used for real-time object classification applications. 

% \subsection{CNN-Based Image classification for Embedded systems}

\section{Model Training} \label{sec:training}
We used a combination of transfer learning and fine-tuning approaches to train the ResNet-50 and Inceptionv3 models. Transfer learning alone on a pre-quantized (int8) trained model with a custom-trained classifier head fails to give appropriate accuracy. Also, transfer learning is not possible on a trained quantized model as it has no trainable parameters. To overcome these issues and use the advantages of a pre-trained model, we use the standard SOTA model (fp32) for transfer learning. It gives us the advantage of pretrained weights instead of random weight initialization which in turn would have required a lot of training.
We replaced the SOTA 1,000 class classifier with a custom classifier head for 9 classes. Model parameters are kept unfrozen with a low learning rate of $1\mathrm{e}{-4}$ for 30 epochs. The entire dataset is divided in a 60:20:20 ratio for training, validation, and testing, similar to that proposed in the original DeepWeeds paper. The models are trained on images with (224,224,3) shape. Most of the other parameters of the training are kept the same as those in the original paper. Adam optimizer \cite{kingma2014adam} and cross-entropy loss function \cite{zhang2018generalized} are used in training. 

\subsection{Network Architecture}

\begin{figure*}[!htbp]
    \centering
    \includegraphics[width=\textwidth]{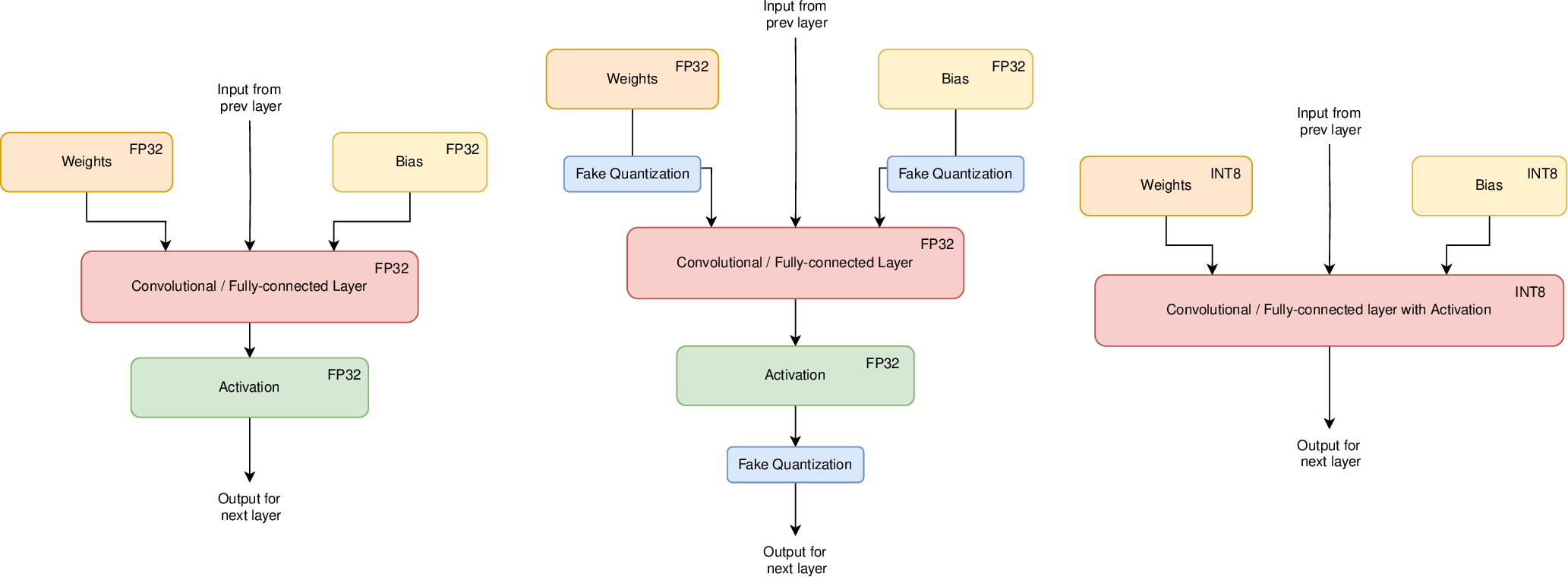}
    \caption{Comparison between full-precision inference (left), model for QAT with simulated quantization (middle), and
quantized model for integer-only inference (right)}
    \label{fig:control}
\end{figure*}

The overall architecture of the feature extractor remains the same for the most part as of a standard ImageNet model, except for the fact that in order to imitate the effects of int8, fake-quantization modules are inserted to model the effects of quantization via zero point shifting and scaling.
To calculate zero point $z$ and scale $s$. 

Consider $x_q \in [\alpha_q, \beta_q]$ and $x \in [\alpha, \beta]$, where $\alpha$ and $\beta$ are minimum and maximum values in their range. 
\begin{equation}
    \begin{split}
    s = \frac{\beta - \alpha}{\beta_q - \alpha_q} \\  
    z = \text{round}\left(\frac{\beta}{\alpha_q} - \frac{\alpha}{\beta_q} - \alpha\right)
    \end{split}
\end{equation}

    where ~$x_q $ = quantized value (int8), ~$x $ = value (fp32), ~$s $ = scale (fp32) and ~$z $ = zero point (int8).

After the training process is completed, model conversion happens, i.e., the activations and weights are quantized to int8 from fp32, and the activations are fused into the preceding layer wherever possible. Since the transition from float to a lesser precision is a lossy process, we typically observe a large decline in accuracy. A quantization-aware training (QAT) \cite{park2018value} is used to assist in reducing this loss.

% \begin{figure}[!htbp]
%     \centering
%     \includegraphics[width=\linewidth]{images/final1.eps}
%     \caption{Original model}
%     \label{fig:control}
% \end{figure}

% \begin{figure}[!htbp]
%     \centering
%     \includegraphics[width=\linewidth]{images/final2.eps}
%     \caption{Modified model for quantization aware training}
%     \label{fig:control}
% \end{figure}

% \begin{figure}[!htbp]
%     \centering
%     \includegraphics[width=\linewidth]{images/final3-3.eps}
%     \caption{Quantized model post training}
%     \label{fig:control}
% \end{figure}

% \begin{figure*}[!htbp]
%     \centering
%     \includegraphics[width=\textwidth]{images/qwid_colored5.eps}
%     \caption{left center right}
%     \label{fig:control}
% \end{figure*}

\subsection{Training Methodology}
In the proposed models, training is done using single-precision floating-point computation. There is no requirement to complete the training in fixed-point because it is done offline on a workstation. Fake-quant modules are inserted to replicate the effects of int8. This technique is termed as quantization-aware training.

\begin{equation}
\begin{split}
    v_{qc} = clip(v_q, \alpha_q, \beta_q) \label{eq:clipping} \\
\text{clip}(x, A, B) =
\begin{cases}
    A & \text{if } x < A \\
    x & \text{if } A \leq x \leq B \\
    B & \text{if } x > B
\end{cases}
\end{split}
\end{equation}
% Clamping: 
% \begin{equation}
%     y_i = min(max(x_i,min value_i),max value_i)
% \end{equation}  

% % \end{math}

QAT is frequently employed with training Q-CNNs and generates results with greater accuracy than static quantization.
 Before the deep neural network is applied to the target, a conversion from a floating-point to a fixed-point representation must be made. This requires quantizing the deep neural network weights because the range of potential values for fixed-point and floating-point representations differs. Non-quantized values continue to be used in the backpropagation. The DNN can be pre-trained using a floating-point representation in order to initialize the parameters with reasonable values. This stabilizes the learning phase with the quantized version and yields better results. Although it is technically possible, there would be extra difficulties with the gradient calculation as discussed in section \ref{sssec:num1}.
\subsubsection{Quantization Mapping}
The mathematical representation of mapping fp32 values to int8 values through quantization:
\begin{equation}
    x_q = round(x/s + z) \label{eq:quant}
\end{equation}
and dequantization:

\begin{equation}
    x = s(x_q - z) \label{eq:dequant}
\end{equation}
% To calculate zero point $z$ and scale $s$ , since $x_q \in [\alpha_q, \beta_q]$ and $x \in [\alpha, \beta]$ where the respective alpha beta is the minimum and maximum values in their range. 
% \begin{equation}
%     \begin{split}
%     s = \frac{\beta - \alpha}{\beta_q - \alpha_q} \\
%     z = \text{round}\left(\frac{\beta}{\alpha_q} - \frac{\alpha}{\beta_q} - \alpha\right)
%     \end{split}
% \end{equation}

%     where ~$x_q $= quantized value, ~$x $= fp32 value, ~$s $= scale which is a floating point value,~$z $= zero point which in an integer value

% \footnote{
%   $v_q$= quantized value, $v$= f32 value, $s$= scale, $z$= zero point
%   $z$ in integer and $s$ is a floating point number
%   }

\subsubsection{Weight Quantization}

CNN-based models are typically formed from convolutional layers and fully connected layers. These do require quantization-aware training for the parameters. The weights of the convolutional layer can be represented in a tensor as 
\begin{math} (f_h, f_w, c_{in}, c_{out}) \end{math},
and for a fully connected layer as
\begin{math} (c_{in}, c_{out}) \end{math} \footnote{
  $c_{\text{in}}$= in channels, $c_{\text{out}}$= out channels, $f_{\text{h}}$= filter height, $f_{\text{w}}$= filter width
  }
The output channel quantization bounds are calculated along each of them.
% \begin{equation}
%     \begin{split}
%     lower bound = min(W) \\
%     upper bound = max(W)
%     \end{split}
% \end{equation} 
% lower-bound = min(W) 
% upper-bound = max(W)
% \end{math}
There may be distinct and independent quantization boundaries for each output channel. This ensures smaller scaling factors and finer quantization ranges than the channel with a higher range in weight. Both Inceptionv3 and ResNet-50 have a large number of weight channels with notable magnitude fluctuations. 
% Layer-wise quantization in this situation would have caused considerable distortion in some of the quantized values.

\subsubsection{Activation Quantization}
The activation functions are quantized by mapping their continuous output values to a discrete range of quantization levels. The numerical precision of activation is decreased during this process, enabling the use of low-precision hardware or memory-efficient deployment. The range of values is calculated similarly to that of convolutional and fully connected layers.

Standard ReLU \cite{agarap2018deep}:

\begin{equation}
    ReLU(x,0,0,1) =
    \begin{cases}
        0 & \text{if } x < 0 \\
        x & \text{if } x \geq 0
    \end{cases}
\end{equation}

Quantized ReLU \cite{leimaoQuantizationNeural}:

\begin{equation}
    ReLU_q(x_q, z_x, z_y, \frac{s_x}{s_y}) =
    \begin{cases}
    z_y & \text{if } x_q < z_x \\
    z_y + \frac{s_x}{s_y} (x_q - z_x) & \text{if } x_q \geq z_x
\end{cases}
\end{equation}
% For some combinations of neural network layers, such as Conv2D-ReLU and Conv2D-BatchNorm-ReLU, layer fusions are frequently used. The idea of several layers is abstracted with layer fusion, which results in a single layer. For instance, in the Conv-ReLU layer, the ReLU function is applied to each component of the receiving field immediately following the application of the Conv filter. We would have the scales and zero points for the inputs \(x0\) to Conv2D, the scales and zero points for the outputs \(y0\) from Conv2D, which are also the inputs to ReLU, and the scales and zero points for the outputs \(y1\) from ReLU without using layer fusion, i.e., we have two separate layers, Conv2D followed by ReLU. The only scales and zero points we would have with layer fusion are those for the inputs \(x0\) to Conv2D-ReLU and those for the outputs \(y0\) from Conv2D-ReLU.
\subsubsection{Layer Fusion}
For some combinations of neural network layers, such as Conv2D-ReLU and Conv2D-BatchNorm-ReLU, layer fusions are frequently used \cite{nagel2021white}. The idea of several layers is abstracted with layer fusion, which results in a single layer. When we do not use layer fusion, meaning we have two separate layers a Conv2D layer followed by a ReLU activation layer we find that we must keep track of scales and zero points at multiple stages within the neural network. Firstly, we need scales and zero points for the inputs, denoted as \(x0\), to the Conv2D layer. Next, we must monitor the scales and zero points for the outputs, \(x1\), coming from the Conv2D layer, which also serves as inputs to the subsequent ReLU activation layer. Lastly, we need to consider the scales and zero points for the outputs, \(x2\), produced by the ReLU activation layer. However, with the introduction of layer fusion, such as combining Conv2D and ReLU into a single Conv2D-ReLU layer, we can simplify this process. In this case, we only need to keep track of scales and zero points for the inputs \(x0\) to the Conv2D-ReLU layer and the outputs \(x1\) emerging from the same Conv2D-ReLU layer. This layer fusion technique streamlines the management of scales and zero points by integrating Conv2D and ReLU into a unified layer, reducing overall complexity. Figure \ref{fig:control} shows fake quantization modules and fused activations.

% sample node from netron explaining this

% \begin{math} 
% quantized value = round((activation value - lower bound) / delta) * delta + lower bound
% \end{math}
\subsubsection{Additional Layers}
The max-pooling layer only uses an element-wise maximum, and there is no need to quantize because the inputs from the preceding layer have already been quantized and the dynamic range cannot be increased. On the other hand, the element-wise addition layer needs quantization as adding two large values of parameters can exceed the dynamic range of the output. Therefore, the output scale factor is calculated using the same quantization method.

\subsubsection{Issues in QAT} \label{sssec:num1}
The inference accuracy from the quantized integer models is invariably worse than that from the floating point models due to information loss. The fact that the floating points are not perfectly recoverable after quantization and dequantization is the cause of this information loss.
\begin{equation}
    x \neq f_{dq}\left(f_q\left(x, s_x, z_x\right), s_x, z_x\right)
\end{equation}
where $f_q$ is quantization function \ref{eq:quant} and $f_{dq}$ is dequantization function \ref{eq:dequant}.

 An error term \(\Delta_x\) is introduced to consider the impact of such information loss during training: 

\begin{equation}
    x = f_{dq}\left(f_q\left(x, s_x, z_x\right), s_x, z_x\right) + \Delta_x
\end{equation}

As a result, the model will have low inference accuracy loss.

\begin{figure}
    \centering
    \includegraphics[width=\linewidth]{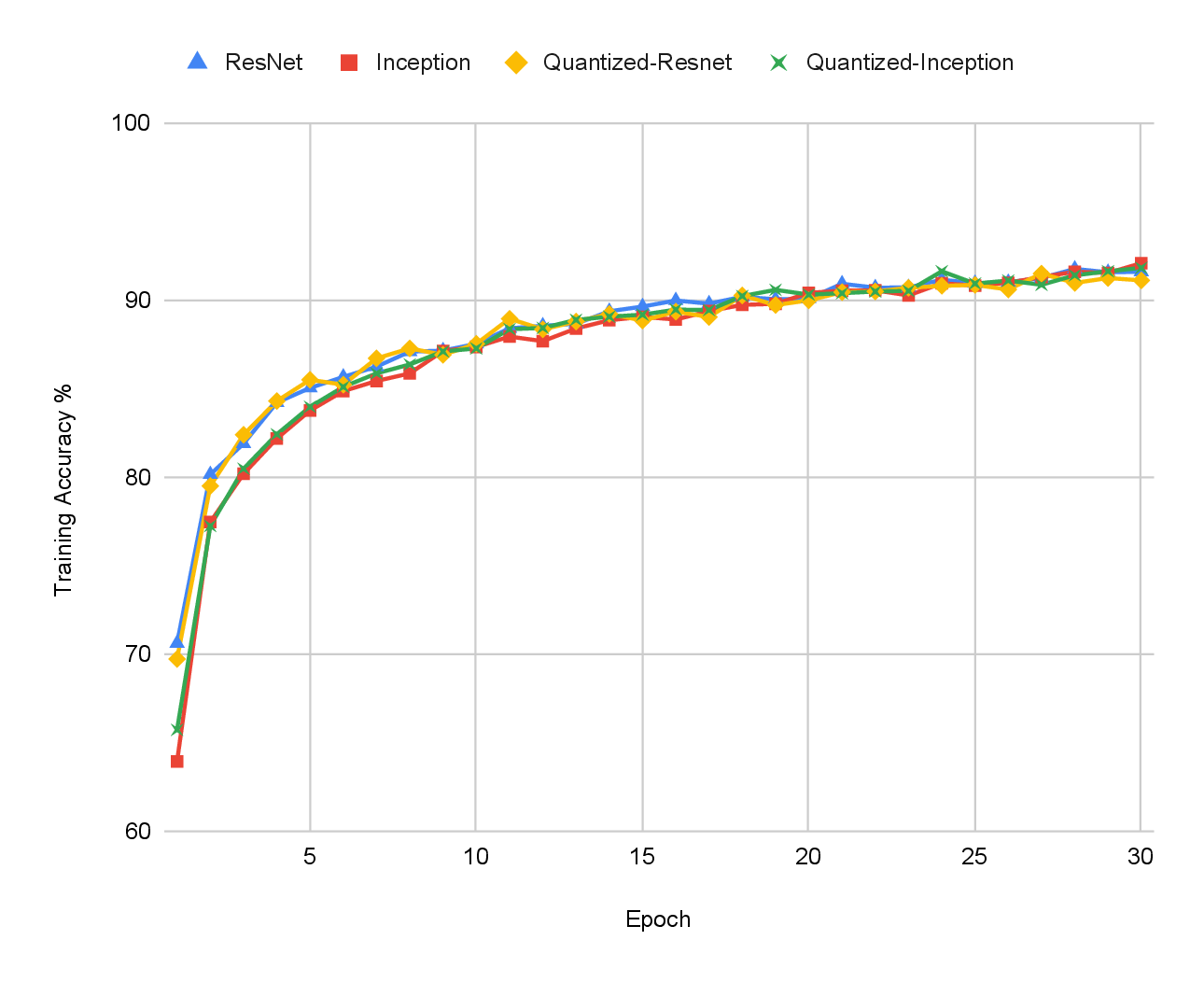}
    \caption{Training Accuracy vs Epochs}
    \label{fig:training}
\end{figure}

Another issue with QAT is that the quantization and dequantization layers are not differentiable. The quantization/dequantization operation maps a continuous input to a discrete output, resulting in a step-like piecewise constant function. This discontinuity in the function makes it non-differentiable at the quantization points. However, there are strategies and procedures that can be used to approximate gradients and make it possible to train quantized networks, such as straight-through estimation \cite{yin2019understanding} and Gumbel-softmax relaxation \cite{NEURIPS2020_90c34175}. These strategies seek to differentiate the quantization layer so that gradient-based optimization is possible while still reaping the benefits of quantization.
% Special care needs to be taken care for batch normalization. During fine tuning, both the batch normalisation settings and the batch statistics are unstable. When the parameters are folded into the previous layer, this significantly increases quantization noise. To overcome this \cite{li2019fully} suggest unfreezing the batchnorm layer during the initial phase then freezing them during the last few epochs in training.
% TODO: add training graphs
% \begin{figure}[!htbp]
%     \centering
%     \includegraphics[width=\linewidth]{images/chart.eps}
%     \caption{Training Accuracy Graph}
%     \label{fig:control}
% \end{figure}
Figure \ref{fig:training} shows the close relation between the SOTA vs their quantized counterparts during training.

\begin{table*}
  \centering
  \begin{threeparttable}
    \caption{Model Comparison}
    \begin{tabular}{>{\centering\arraybackslash}ccccccc}
      \toprule
      \textbf{Model Name} & \multicolumn{2}{c}{\textbf{Accuracy \%}} & \textbf{Model Size (MB)} & \multicolumn{3}{c}{\textbf{Inference Time (ms)}} \\
      \cmidrule(lr){2-3} \cmidrule(lr){5-7} 
      & \textbf{Top1} & \textbf{Top3} & & \textbf{Core-i5} & \textbf{Tensor-G2} & \textbf{Cortex-A72} \\
      \midrule
      ResNet-50 & 95.95 & 99.43 & 94.45 & 112.71 & 173.62 & 1906.73\\
      Quantized ResNet-50 & 94.77 & 99.49 & 23.72 & 42.61 & 90.99 & 218.42\\ 
      Inceptionv3 & 95.09 & 99.63 & 87.59 & 90.17 & 120.92 & 838.79\\
      Quantized Inceptionv3 & 94.57 & 99.40 & 22.04 & 34.46 & 71.15 & 187.56 \\ 
      \bottomrule
      \label{tab:modelcompare}
    \end{tabular}
    \begin{tablenotes}
      \item[] Note: Accuracy values represent the Top1 and Top3 metrics. Model Size is in megabytes (MB), and Inference Time is measured in milliseconds (ms) on different hardware platforms.
    \end{tablenotes}
  \end{threeparttable}
  % \label{tab:modelcompare}
\end{table*}

\section{Results} \label{sec:results}
The training of the proposed models was conducted on Nvidia Tesla P100 GPU and Intel Xeon 2.20 GHz CPU. The programming environment used for training was Python 3.9 and the deep learning framework was PyTorch 2.0.1. The inference was made on three different hardware, i.e., PC with Intel Core i5-8250U from the x86 family, and for the ARM architecture we are using a Mobile device with Tensor G2, and Raspberry Pi with Cortex-A72 for a complete sense of inference time on different architecture coverage. The hardware specifications are presented in Table \ref{tab:harwarecompare}. The input image was run for 100 iterations (forward pass only) to get the average iterations per second shown in Table \ref{tab:modelcompare}. The inference time does not include the pre-processing of the input-image tensor nor the model loading time. In Table \ref{tab:performancecompare}, we compare the respective total operations, memory footprint represents the maximum RAM usage by any layer for a single image during inference.

\begin{figure}[H]
    \centering
    \includegraphics[width=\linewidth]{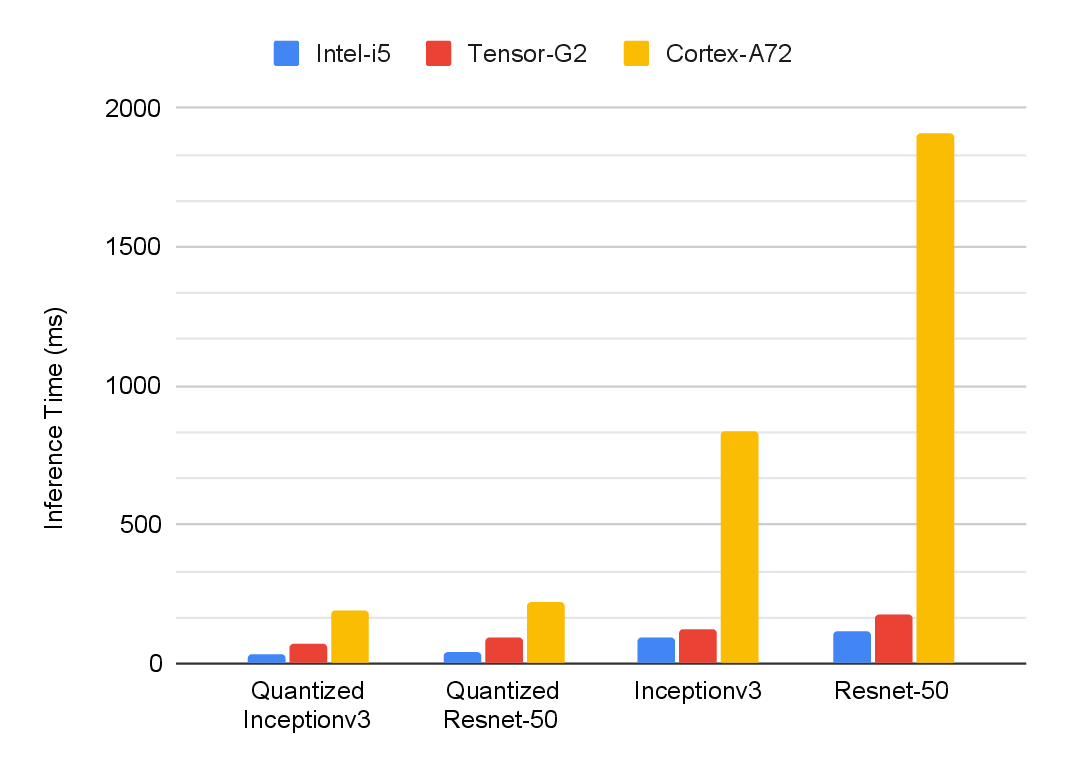}
    \caption{Inference Time Comparison}
    \label{fig:inferencecomp}
\end{figure}

\begin{figure*}[!htbp]
    \centering
    \includegraphics[width=\textwidth]{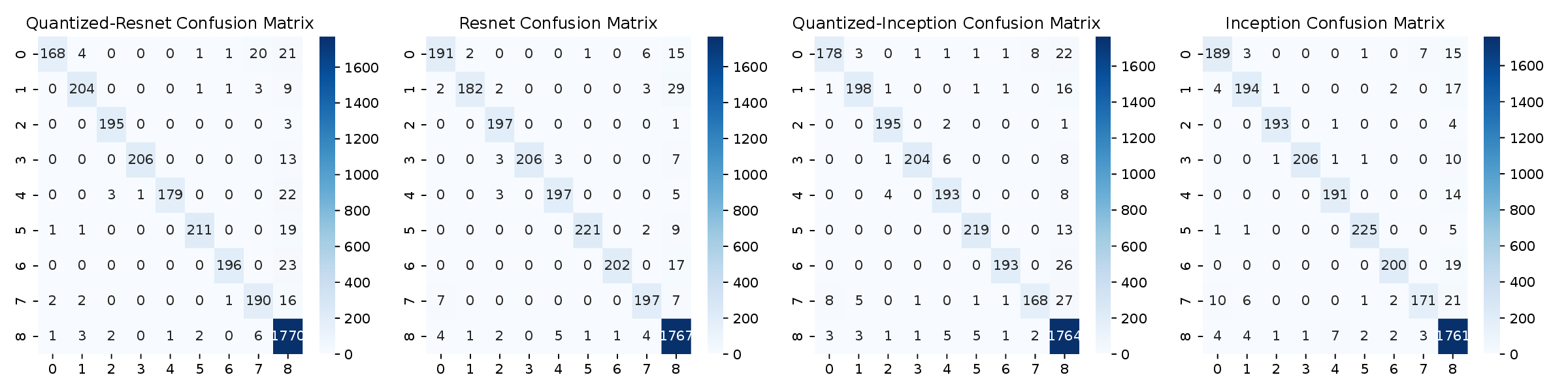}
    \caption{Confusion matrix of respective models, X-axis represents the actual label, and Y-axis represents the predicted label. The respective labels are in the following sequence: 0: 'Chinese apple', 1: 'Lantana', 2: 'Parkinsonia', 3: 'Parthenium', 4: 'Prickly acacia', 5: 'Rubber vine', 6: 'Siam weed', 7: 'Snake weed', 8: 'Negative'.}
    \label{fig:conf}
\end{figure*}

Also, accuracy cannot be a single tool to evaluate the model results because of the presence of unbalanced classes. So, a confusion matrix provides a much clearer picture of the trained model for each class. In Figure \ref{fig:conf}, we can see both SOTA models and their quantized counterparts perform equally well. Due to the uneven sample distribution, there are more instances where the eighth category or other categories are incorrectly identified. However, this inaccuracy rate can be tolerated because the negative weed category has approximately eight times as many images as the other categories. In Table \ref{tab:through} we further represent our findings with model throughput on different hardware devices.

% \textbf{Accuracy}: Top1,top5, roc curves/confusion matrix for both resnet and inception 
% \textbf{Model Size}:
% \textbf{Inference time}:

% \begin{table}[htbp]
%     \centering
%     \caption{Model Comparison}
%     \begin{tabular*}{\textwidth}{@{\extracolsep{\fill}}|X|X|X|X|X|}
%         \hline
%         Model Name & Top1 & Top5 & Model Size (MB) & Inference Time (ms) \\
%         \hline
%         ResNet & & & & \\
%         \hline
%         ResNet Quantized & & & & \\
%         \hline
%         Inceptionv3 & & & & \\
%         \hline
%         Inceptionv3 Quantized & & & & \\
%         \hline
%     \end{tabular*}
% \end{table}

% \usepackage{tabularx} % in the preamble
% ....

\begin{table}[H]
  \centering
  \begin{threeparttable}
    \caption{Performance Analysis}
    \begin{tabular}{ccc}
      \toprule
      \label{tab:performancecompare}
      \textbf{Model Name} & \textbf{GFLOPs/GOPs} & \textbf{Memory footprint (Mb)} \\
      \midrule
      ResNet-50 & 4.13 & 53.60\\
      Quantized ResNet-50 & 4.13 & 6.70\\ 
      Inceptionv3 & 2.85 & 34.64\\
      Quantized Inceptionv3 & 2.85 & 4.58\\ 
      \bottomrule
    \end{tabular}
    \begin{tablenotes}
      \item Note: The complexity of the Non-Quantized models are represented in floating-point operations while quantized models use operations.
    \end{tablenotes}
  \end{threeparttable}
\end{table}

\begin{table}[H]
  \centering
  \begin{threeparttable}
    \caption{Hardware Specification}
    \begin{tabular}{cc@{\hskip 0.05in}c@{\hskip 0.05in}c}
      \toprule
      \label{tab:harwarecompare}
      \textbf{Hardware} & \textbf{Memory(GB)} & \textbf{Clock Speed(GHz)} & \textbf{CPU-Cores/Threads}\\
      \midrule
      Core-i5 & 8 & 3.40 & 4/8\\
      Tensor-G2 & 12 & 2.85 & 8/8\\ 
      Cortex-A72 & 8 & 1.80 & 4/4\\
      \bottomrule
    \end{tabular}
    \begin{tablenotes}
      \item Note: Different processors used are compared for clarity in performance capabilities. 
    \end{tablenotes}
  \end{threeparttable}
\end{table}

\begin{table*}[h]
  \centering
  \begin{threeparttable}
    \caption{Throughput Comparision}
    \begin{tabular}{ccccc}
      \toprule
      \label{tab:through}
      \textbf{Hardware} & \multicolumn{4}{c}{\textbf{Model Throughput GFLOPs/GOPs per sec}}\\
      \cmidrule(lr){2-5} & {\textbf{ResNet-50}} & \textbf{Quantized ResNet-50} & \textbf{Inceptionv3} & \textbf{Quantized Inceptionv3} \\
      \midrule
      Tensor-G2 & 23.79 & 45.40 & 23.60 & 40.12\\
      Cortex-A72 & 2.16 & 18.91 & 3.40 & 15.21 \\ 
      Core-i5 & 36.65 & 96.96 & 31.65 & 82.83\\
      \bottomrule
    \end{tabular}
    \begin{tablenotes}
      \item Note: Comparing throughput of models with their quantized counterparts on the selected hardware.
    \end{tablenotes}
  \end{threeparttable}
\end{table*}
\section{Conclusion} \label{sec:conclusion}
In this paper, we proposed a Quantized ResNet-50 and Inceptionv3 model, a low complexity fully convolutional neural network %integer-only arithmetic inference, image recognition models%
for non-GPU enabled and embedded devices. These models perform comparably with respect to their SOTA counterparts in terms of accuracy. In terms of storage consumption, it takes almost 4$\times$ less storage, achieving a 4$\times$ speedup on CPU, 6$\times$ speedup on Raspberry Pi, and 2$\times$ speedup on a mobile processor. We believe that this strategy and the findings from our experimental study will make it easier to conduct future quantization research and develop industrial vision applications tailored to agriculture for devices with limited resources, enabling them to be more AI-enabled while using fewer resources.
 \newpage

\bibliography{peRef}
\bibliographystyle{ieeetr}

\end{document}